# Discovering shared and individual latent structure in multiple time series

**Suchi Saria**[1], **Daphne Koller**[1] **and Anna Penn**[2]
[1]Department of Computer Science, [2]Department of Pediatrics
Stanford University
Stanford, CA 94305
ssaria@cs.stanford.edu

## Abstract

This paper proposes a nonparametric Bayesian method for exploratory data analysis and feature construction in continuous time series. Our method focuses on understanding shared features in a set of time series that exhibit significant individual variability. Our method builds on the framework of latent Diricihlet allocation (LDA) and its extension to hierarchical Dirichlet processes, which allows us to characterize each series as switching between latent "topics", where each topic is characterized as a distribution over "words" that specify the series dynamics. However, unlike standard applications of LDA, we discover the words as we learn the model. We apply this model to the task of tracking the physiological signals of premature infants; our model obtains clinically significant insights as well as useful features for supervised learning tasks.

## 1 Introduction

Time series data is ubiquitous. The task of knowledge discovery from such data is useful in many scientific disciplines including understanding disease pathogenesis from longitudinal studies, mining social interactions, and wildlife monitoring. Dynamic models that can succinctly capture generation of measured data as a function of underlying latent states can serve as a lens into understanding series structure. For example, generation of physiologic data by latent disease states can help reveal how the set of diseases manifest and uncover novel disease associations.

Traditional generative models for time series data (such as switching Kalman filters[1]) do not explicitly model generation of a continuum of heterogeneous exemplar series. However, in the domain of modeling disease physiology, the set of clinical diagnoses for most diseases is not naturally discrete; patients suffer from diseases to varying extents, and even within disease labels, patients exhibit significant variability. Furthermore, often the diagnosis is the result of a clinical observation with significant uncertainty regarding both the onset and the nature of the disease. Increasing representation granularity by increasing the number of classes can help, but ad hoc discretization into a fixed set limits our ability to model instance-specific variability. Moreover, the combinatorics quickly get out of hand when one considers combinations of different diseases, a situation that is unfortunately quite common in practice.

In the domain of natural language processing, topic models have found great success as a representation for uncovering the underlying structure of document corpora [2, 7]. A document is associated with a distribution over latent variables called *topics*, each of which defines a distribution over words. At a high level, this paradigm extends naturally to applications such as disease modeling: individual patients maintain their own distributions over disease topics, so each patient can be expressed as a point lying in the continuous disease (topic) simplex. Each topic defines a distribution over dynamic behaviors observed in the time series. These dynamic behaviors, play the role of "words" in this framework; these words are shared across topics, allowing us to uncover relationships between different disease states.

To construct this approach, we build on the framework of *hierarchical Dirichlet processes*, which was designed to allow sharing of mixture components within a multi-level hierarchy. While conceptually straightforward, this extension involves some significant subtleties. Most obviously, current implementations of topic models assume that the expressed features are pre-specified and only the latent topic variables are inferred. To apply this framework to continuous data such as physiologic time series, we must define a notion of a "word". Loosely speaking, a word specifies the parameterization of the function that generates the system dynamics for the duration of the word. Moreover, the duration of the word is also not fixed in advance, and may vary from one occurrence to the next. Hence, our representation of a time series in terms of words must also postulate word boundaries.

In the remainder of this paper, we define our generative *time series topic model* (TSTM), and provide an efficient block Gibbs sampling scheme for performing full Bayesian inference over the model. We then present results on our target application of analyzing physiological time series data, demonstrating its usefulness both for analyzing the behavior of different time series, and for constructing features that are subsequently useful in a supervised learning task.

## 2 Related Work

An enormous body of work has been devoted to the task of modeling time series data. Probabilistic generative models, the category to which our work belongs, typically utilize a variant of a switching dynamical system [1], where one or more discrete state variables determine the momentary system dynamics, which can be either linear or in a richer parametric class. However, these methods typically utilize a single model for all the time series in the data set, or at most define a mixture over such models, using a limited set of classes. These methods are therefore unable to capture significant individual variations in the dynamics of the trajectories for different patients, as required in our data.

Recent work by Fox and colleagues [5, 3, 4] uses nonparametric Bayesian models for capturing generation of continuous-valued time series. Conceptually, the present work is most closely related to BP-AR-HMMs [5], which use Beta processes to share observation models, characterized by autoregressive processes (AR), across series. Thus, it captures variability between series by sampling subsets of low-level features that are specific to individual series. However, BP-AR-HMMs are aimed at capturing individual variation that manifests as words (features) that are unique to a particular time series. By comparison, we aim to capture higher-level concepts that occur broadly across a subpopulation (e.g., physiologic characteristics exhibited by *Lung disorders*) while modeling series-specific variability in terms of the extent to which these higher-level concepts are expressed (e.g., transient respiratory disorder versus long term respirator distress). Other works [3, 4] have utilized hierarchical Dirichlet process priors for inferring the number of features in hidden Markov models and switching linear dynamical systems, but these models do not try to represent variability across exemplar series. Temporal extensions of LDA [19, 20] model evolution of topic compositions over time in text data but not continuous-valued temporal data.

A very different approach to analyzing time series data is to attempt to extract meaningful features from the trajectory without necessarily constructing a generative model. For example, one standard procedure is to re-encode the time series using a Fourier or wavelet basis, and then look for large coefficients in this representation. However, the resulting coefficients do not capture signals that are meaningful in the space of the original signal, and are therefore hard to interpret. Features can also be constructed using alternative methods that produce more interpretable output, such as the work on sparse bases [11]. However, this class of methods, as well as others [13], require that we first select a window length for identifying common features, whereas no such natural granularity exists in many applications. Moreover, none of these methods aim to discover higher level structure where words are associated with different "topics" to different extents.

Our specific domain of neonatal monitoring has been previously studied in a machine learning setting [21, 14], but focusing on the different task of monitoring for known events (e.g., sensor drops).

## 3 Time Series Topic Model

Time Series Topic Model (TSTM) is an extension of Latent Dirichlet Allocation (LDA) [2] for time-series data. Latent Dirichlet Allocation (LDA) is a generative probabilistic model for text corpora. It makes the assumption that there is an underlying fixed set of topics that is common to the heterogeneous collection of documents in the corpus. A topic is a distribution over the vocabulary of all words present in the corpus. A document is generated by first choosing a document-specific mixing proportion over the topics. To sample each word, sample a topic, and then sample a word from that topics' distribution over words.

Unlike text, in time series data often the features to be extracted are structurally not obvious (see figure 2). Pre-segmenting the sequence data into "words" does not offer sufficient flexibility to learn from the data, especially in the realm of exploration for knowledge discovery. Thus, we integrate feature discovery into a temporal extension of LDA. We describe below each of the TSTM components (in a bottom-up fashion).

**Data generation model:** We assume that the continuous-valued data $y_t$ at time $t$ is generated using a function $f_k$ from a finite set $\mathcal{F}$ of generating functions. These functions take as inputs $x_t$, values dependent on current and previous time slices, and generate the output as $y_t = f_k(x_t; \theta_k)$. $f_k$, an expressive characterization of the time series dynamics, can be thought of as th $kth$ word in the time-series corpus vocabulary. Hereafter, we use *words* and *generating functions* interchangeably. The parameterization of $f_k$ depends on the choice of the observation model. We choose to use vector autoregressive processes, which are used for temporal modeling in numerous domains, including medical time series of fMRI and EEG. Other observation models (such as an SLDS with mixture model emissions [3]) can also be used.

In an order $p$ autoregressive process, given a function $f_k$ with parameters $\{A^k, V^k\}$, the observation is generated as:

$$\vec{y}_t = A^k X_t^T + \vec{v}_t \qquad \vec{v}_t \sim \mathcal{N}(0, V^k) \qquad (1)$$

where $\vec{y}_t \in \mathcal{R}^{m \times 1}$ for an m-dimensional series. The inputs $X_t = [\vec{y}_{t-1}, \ldots, \vec{y}_{t-p}]$. Parameters $A^k \in \mathcal{R}^{m \times p}$, and $V^k$ is an $m \times m$ positive-semidefinite covariance matrix. The k$th$ word then corresponds to a specific instantiation of the generating function parameters $\{A^k, V^k\}$. For TSTM, we want the words to persist for more than one time-step. Thus, for each word, we have an additional parameter $\omega_k$ that specifies mean length of the word as $1/\omega_k$. Our goal now is to uncover this set of functions $\mathcal{F}$, via instantiations of the generating function parameters, denoted more generally as $\vec{\theta}_k \in \Theta$.

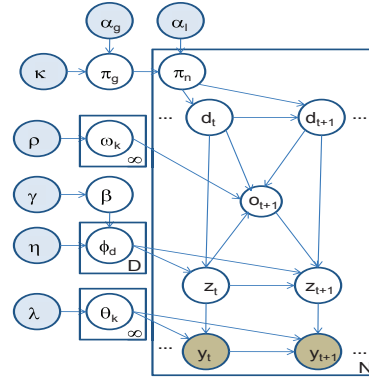

Figure 1: Graphical representation of the time series topic model

**Word and Topic descriptions:** To uncover the finite generating function set $\mathcal{F}$, such that these functions are shared across latent topics, we use the hierarchical Dirichlet process[18] (HDP). A Dirichlet process (DP), denoted by $\text{DP}(\gamma, H)$, yields a distribution on discrete measures. $H$ is a base (continuous or discrete) probability measure on $\Theta$ and $\gamma$ is the concentration parameter. Sethuraman [17] shows that $G_0 \sim \text{DP}(\gamma, H)$, a sample drawn from the DP, is a discrete distribution because, with probability one:

$$G_0 = \Sigma_{k=1}^{\infty} \beta_k \delta_{\theta_k} \qquad \theta_k \sim H \qquad \beta_k = \beta_k' \prod_{l=1}^{k-1}(1 - \beta_l') \qquad \beta_k' \sim \text{Beta}(1, \gamma) \qquad (2)$$

Draws from $H$ yield the location of "sticks"', $\delta_{\theta_k}$, in the discrete distribution. The generation of the stick weights $\beta_k$ is denoted by $\vec{\beta} \sim \text{GEM}(\gamma)$. Thus, we use the DP to place priors on the mixture of generating functions by associating a generating function with each stick in $G_0$. In addition, by associating each data sample (time point in a series) with a specific generating function via indicator variables, the posterior distribution yields a probability distribution on different partitions of the data. The mixing proportion (the stick weights) in the posterior distribution are obtained from aggregating corresponding weights from the prior and the assigned data samples.

HDPs [18] extend the DP to enable sharing of generating functions between topics. If discrete measures $G_j$ are sampled with the discrete measure $G_0$ as its base measure, the resulting distributions have non-zero probability of regenerating the same sticks, thereby sharing generating functions between related topics. Thus, a draw $G_j$ from the HDP with $G_0$ as its base measure, $G_d \sim \text{DP}(\eta, G_o)$, can be described as:

$$G_d = \Sigma_{k=1}^{\infty} \phi_{dk} \delta_{\theta_k} \qquad \phi_d \sim DP(\eta, \beta), \quad \beta \sim GEM(\gamma), \quad \theta_k \sim H \qquad (3)$$

where $\phi_d$ represents the topic specific mixing proportions over the generating functions and $\beta$ represents the global mixing proportion. Similar to [5], we use a matrix-normal inverse-Wishart on the parameters $\{A^k, V^k\}$ and a symmetric Beta prior on $\omega_k$ as our base measures $H$.

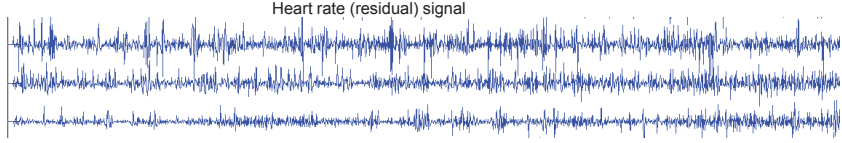

Figure 2: Heart signal (mean removed) from three infants in their first few hours of life

**Dynamics of words and topics:** Given the words $\mathcal{F}$, topics $\phi_{1:D}$ and series-specific transition matrices $\pi_n$, the series generation is straightforward. For each time slice $t \in 1, \cdots, T$, we generate:

1. Current latent topic state given topic at previous time-step, $d_t \sim \text{Mult}(\pi_n^{d_{t-1}})$
2. Switching variables $o_t$, which determine whether a new word is selected. A new word is always generated ($o_t = 0$) if the latent state has changed from the previous time step; otherwise, $o_t$ is selected from a Bernoulli distribution whose parameter determines the word length. Thus, $o_t \sim I(d_t = d_{t-1})\text{Bernoulli}(\omega_{z_{t-1}})$, where $I$ is the indicator function.
3. The identity of the generating function (word) to be applied; if $o_t = 1$, we have $z_t = z_{t-1}$, otherwise $z_t \sim \text{Mult}(\phi_{d_t})$.
4. Observation given the generating function index $z_t$ as $y_t \sim f_{z_t}(x_{t-1}; \theta_{z_t})$.

The-series specific topic transition distribution $\pi_n$ is generated from the global topic transition distribution $\pi_g$. Hyperparameters $\alpha_l$ controls the degree of sharing across series in our belief about the prevalence of latent topic states. A large $\alpha_l$ assigns a stronger prior and allows less variability across series. To generate $\pi_n$, each row $i$ is generated from $\text{Dir}(\alpha_l \pi_g^i)$, where $\pi_g^i$ is the *ith* row of the global topic transition distribution. Given hyperparameters, $\alpha_g$ and $\kappa$, $\pi_g^i \sim Dir(\alpha_g + \kappa \delta_i)$. $\kappa$ controls the degree of self-transitions for the individual topics.

## 4 Approximate Inference using block-Gibbs

Several approximate inference algorithms have been developed for mixture modeling using the HDP; see [18, 3, 10] for a discussion and comparison. We use a block-Gibbs sampler that relies on the *degree L weak limit* approximation presented in [8]. This sampler has the dvantage of being simple, computationally efficient and shows faster mixing than most alternate sampling schemes [3].

The block-Gibbs sampler for TSTM proceeds by alternating between sampling of the state variables $\{d_t, z_t\}$, the model parameters, and the series specific transition matrices. To introduce notation briefly, we use $1 : T$ to mean all indices 1 through $T$. $n$ indexes individual series. We drop sub-indices when all instances of a variable are used (for e.g., $z_{1:N,1:T_n}$ is written as $z$ for short). We drop the index $n$ when explicit that that the variable refers to a single series. We detail the update steps of our block-Gibbs inference algorithm below.

**Sampling latent topic descriptions $\beta$, $\phi_d$:** The DP can also be viewed as the infinite limit of the order $L$ mixture model [8, 18]:

$$\beta|\gamma \sim \text{Dirichlet}(\gamma/L, \cdots, \gamma/L) \qquad \phi_d \sim \text{Dirichlet}(\eta\beta) \qquad \theta_k \sim H \qquad (4)$$

We can approximate the limit by choosing $L$ to be larger than the expected number of words in the data set. The prior distribution over each topic-specific word distribution is then:

$$\phi_d|\beta, \eta \sim \text{Dir}(\eta\beta_1, \cdots, \eta\beta_L) \qquad (5)$$

Within an iteration of the sampler, let $m_{d,l}$ be the counts for the number of times $z_{n,t}$ sampled the $l$th word[1] for the $d$th disease topic; that is, let $m_{d,l} = \sum_{n=1:N} \sum_{t=1:T_n} I(z_{n,t} = l) I(d_{n,t} = d) I(o_{n,t} = 0)$ and $m_{\cdot,l} = \sum_{d=1:D} m_{d,l}$. The posterior distribution for the global and individual topic parameters is:

$$\beta | z, d, \gamma \sim \text{Dir}(\gamma/L + m_{\cdot,1}, \cdots, \gamma/L + m_{\cdot,L}) \tag{6}$$

$$\phi_{d'} | z, d, \eta, \beta \sim \text{Dir}(\eta\beta_1 + m_{d',1}, \cdots, \eta\beta_L + m_{d',L}) \tag{7}$$

**Sampling word parameters $\omega_l$ and $\theta_l$:** Loosely, the mean word length of the $l$th word is $1/\omega_l$. A symmetric Beta prior with hyperparameter $\rho$, conjugate to the Bernoulli distribution, is used as a prior over word lengths. The sufficient statistics needed for the posterior distribution of $\omega_l$ are the counts $\bar{c}_{l,i} = \sum_{n=1:N} \sum_{t=1:T_n} I(d_{n,t} = d_{n,t-1}) I(z_{n,t-1} = l) I(o_{n,t} = i)$ where $i \in \{0, 1\}$, representing the number of time steps, across all sequences, in which the topic remained the same, the word was initially $l$, and the word either changed ($o_{n,t} = 1$) or not ($o_{n,t} = 0$). Thus,

$$\omega_l | \bar{c}_{l,\cdot}, \rho \sim \text{Beta}(\rho/2 + \bar{c}_{l,1}, \rho/2 + \bar{c}_{l,0}) \tag{8}$$

For sampling the AR generating function parameters, note that conditioned on the mode assignments $z$, the observations $y_{1:T,1:N}$ can be partitioned into sets corresponding to each unique $l \in L$. This gives rise to $L$ independent linear regression problems of the form $Y^l = A^l X^l + E^l$ where $Y^l$ is the target variable, with observations generated from mode $l$, stacked column-wise. $X^l$ is a matrix with the corresponding $r$ lagged observations and $E^l$ is the corresponding noise matrix. The parameters $A^l$ and $V^l$ are sampled from the posterior given conjugate priors of the Matrix-Normal Inverse-Wishart, similar to [5].

**Sampling global and series-specific transition matrices, $\pi_g$ and $\pi_n$:** Since the number of topic states $D$ is known, and we use conjugate priors of Dirichlet distribution for each row of the transition matrix, the posterior update simply involves summing up counts from the prior and the data. The relevant count vectors are computed as $c_{n,k}^i = \sum_{t=1}^{T_n} I(d_{n,t-1} = i) I(d_{n,t} = k)$ and $c_k^i = \sum_{n=1}^{N} c_{n,k}^i$ which aggregates over each series. $\vec{c^i} = \{c_1^i, \cdots, c_D^i\}$ and $i$ indexes a row of the transition matrix:

$$\pi_g^i | d, \alpha_g, \kappa \sim \text{Dir}(\alpha_g + \kappa\delta_i + \vec{c^i}) \qquad \pi_n^i | \pi_g, d, \alpha_l \sim \text{Dir}(\alpha_l \pi_g^i + c_{n,1:D}^i) \tag{9}$$

**Sampling state variables:** If all model parameters (topic and word descriptions) are specified, then one can exploit the structure of the dependency graph to compute the posterior over the state variables using a single forward-backward pass. This is the key motivation behind using block Gibbs. The joint posterior can be computed recursively. Forward sampling is used to sample the variables in each time slice given samples for the previous time slice:

$$P(z_{1:T}, d_{1:T} | y_{1:T}, \vec{\pi}) = \prod_t P(z_t, d_t | z_{t-1}, d_{t-1}, y_{1:T}, \vec{\pi}) \tag{10}$$

Top-down sampling is used within a given time slice. See appendix A for details.

## 5 Experiments and Results

We demonstrate the utility of TSTM on physiologic heart rate (HR) signal collected from premature infants of gestational age $\leq 34$ weeks and birth weight $\leq 2000$ grams admitted to the XXX (anonymized) neonatal ICU within the first few hours of life. Our inclusion criteria resulted in data from 145 infants.[2] These infants are continuously monitored as part of routine care and data aggregated at the minute-level granularity was stored. These infants are extremely vulnerable, and complications during their stay in the NICU can adversely affect long term neurological development. Clinicians and alert systems implemented on ICU monitors utilize coarse information based on thresholding of the signal mean (e.g., HR $> 160$ bpm) and discard the dynamics as noise. We use TSTM to discover whether there is information contained in the signal dynamics. As infants mature, they become less vulnerable; their base heart rate also lowers. We preprocess the signal to remove the *basal* signal computed from taking a 40 minute moving average window. Removing the base signal allows us to capture characteristics only related to the dynamics of the signal (see raw signal in figure 2).

---

[1] Within this approximation, words get re-ordered such that all words that are observed in the corpus are assigned indices less than $L$. Thus, $l$ indexes the $l$th observed word which can correspond to different parameter instantiations over different iterations.

[2] All work was performed under protocol 8312 approved by XXX's (anonymized) Panel on Human Subjects.

**Partially-supervised training:** We first experimented with a partially-supervised training regime of labeled LDA [15], where during training, we constrain infant-specific transition matrices *not* to have topics corresponding to complications that they did not show symptoms for. This form of training has the advantage of biasing the topics into categories that are coherent and more easily interpreted. We define four broad clinically meaningful topics: *Lung* for primarily lung related complications such as RDS; *Hemorrhage* for head related complications such as IVH; *Multi* as the catch-all class for severe complications that often affect multiple organ systems; and *Healthy*.

For each infant in a randomly chosen subset of 30 infants, we assign a vector $\lambda_n$ of length $D = 4$, where we have a 0 at index $i$ when this infant is known not to have complications related to the *ith* category. All infants are marked to have the healthy topic, representing the assumption that there is some fraction of their time in the NICU at which they exhibit behaviors of a healthy baby.

Each row of the infant-specific transition matrix is generated as:

$$\pi_n^i \sim \text{Dir}\left(\alpha_l \frac{\pi_g^i \otimes \lambda_n}{<\pi_g^i, \lambda_n>}\right) \quad \lambda_n(i) = 1 \tag{11}$$

where $\otimes$ denotes the element-wise vector product. We used an AR(1) process as our observation model and set the $L$ to 15. We experimented with a few different settings of the hyperparameters; upon visual inspection, the relative distribution of words for each infant looked similar. For the reported experiments, $\alpha_l$, $\gamma$ and $\eta$ were each set to 10, $\kappa = 25$ and $\rho = 20$. We run 2000 iterations for each Gibbs chain.

**Qualitative evaluation:** We fix the topic distributions $\phi_{1:D}$ to that of the 2000*th* Gibbs iteration and run inference on our larger set of 145 infants. Here, no supervision is given; that is, both $\pi_g$ and $\pi_n$ are initialized from the prior and are left unconstrained during the inference process (using block Gibbs). We ran three separate Gibbs chains to 400 iterations. Given the topics, the block-Gibbs sampler appears to mix within 200 iterations.

In figure 3, we analyze 30 randomly selected infants from this test set at the 400*th* iteration from chain 1. In panel (a), we plot the word distribution for days 1,2 (top) and days 7,8 (bottom). Several interesting observations arise: the infants follow a continuum of word distribution profiles. Infants with no complications, shown as red squares at the bottom of panel (a), express words 3, 9 and 10 in much greater proportion than the other infants. As can be seen in (b), these words are primarily associated with the *Healthy* topic. Figure 3d shows examples of words extracted from the data. The words with the highest mutual information with the *Healthy* topic appear to have higher frequencies. Words with AR parameter $a > 1$ generally represent heart rate accelerations (e.g., word 8 shown in gray); words where $a$ is positive and close to 0 represent periods with significantly lower dynamic range (e.g., word 2 shown in purple). Thus, our model seems to imply a correlation between the health of the infant and the extent to which their heart rate exhibits significant variations.

Respiratory distress (RDS), a common complication of prematurity, usually resolves within the first few days as the infant stabilizes and is transitioned to room air. This is reflected by the decrease in relative proportion of word 2, only associated with the *Lung* topic. Exceptions to this are infants 3 and 30, both of whom have chronic lung problems.

To demonstrate the insights provided by this representation, we examine the clinical evolution of three sample infants — 2, 16 and 23 — chosen to be illustrative of different trajectories of the word distributions. In figure 3c, we show the posterior for the *Healthy* topic being expressed from averaging over 30 Gibbs chains. The bold line shows the smoothed posterior over time. Infant 2 (I2) was born with a heart defect called VSD (can be acutely life-threatening) and a moderate size patent ductus arteriosus (PDA), both of which cause left to right shunting and disrupt blood flow to the body. I2 was ligated on day 4, a procedure performed to resolve PDA. She was also on dopamine starting day 2 due to hypotension, had a renal failure on day 3 post indomethacine (medication for PDA) and was on a ventilator during this entire time. On Day 7, her state started to resolve significantly, and on day 8 her ventilator settings were minimal and she was taken off dopamine. Her empirical evolution closely tracks her medical history; in particular, her state continually improves after day 4. Infant 16 was a healthier preemie with few complications of prematurity and was discharged on day 4. Infant 23, on the other hand, got progressively sicker and eventually died on day 4. The figure shows that their inferred posterior prediction closely tracks their medical history as well. Of note, the ventilator controls only oxygen supply and does not directly control the heart rate, so the inferred topics are not just uncovering the different ventilator settings.

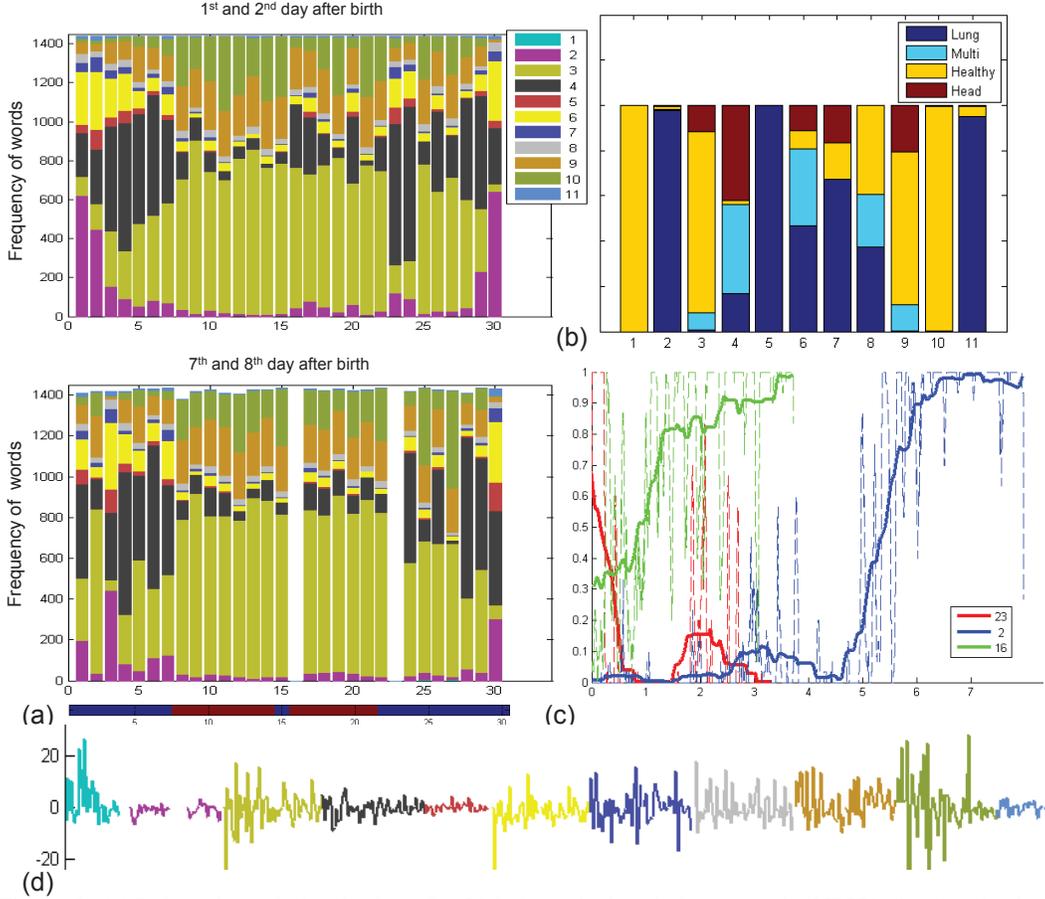

Figure 3: (a) Inferred word distributions for 30 infants during their stay at the NICU, (b) distribution over disease topic given words for the population, (c) inferred posterior (and smoothed version shown in bold) over latent state, *Healthy*, (d) examples of inferred features extracted from the data.

**Quantitative evaluation:** To provide a more quantitative evaluation, we extract grades $G_{1:N}$ that represent an infant's health based on his final outcome, as identified retrospectively by a clinician. Grade 0 represents no complications; grade 1 to isolated minor complications frequently associated with prematurity; grade 2 to multiple minor complications; grade 3 to major complications at low grade; and grade 5 to major complications at severe grades. We defined a "health ranking" $H$ over infants, based on their relative frequency of the *Healthy* topic. The rank score for a ranking $H$ is:

$$\text{rankscore}_H = \Sigma_{n=1}^{N} \Sigma_{m=1}^{N} I(H(n) > H(m))(G_n - G_m) \qquad (12)$$

The maximal achievable score for our dataset is 22992 for the perfect ordering while a random ordering has an expected score of 0. For comparison, we compute the rank score from SNAP [16], a validated and highly accurate score that characterizes illness severity based on birth characteristics, laboratory tests and monitor data at 12 hours of life. SNAP achieves a score of 13030, corresponding to 78.3% of the overall range. Based on dynamics alone, TSTM (in the median from three Gibbs trials) achieves a score of 10432 corresponding to 72.69% at the first 12 hours and 11314 corresponding to 74.6% from the first 8 days. Although SNAP achieves a slightly higher score, it incorporates multiple biomarkers including invasive tests.

**Comparison to Fourier features:** The preceding analysis ranked babies using only a single feature extracted from the TSTM model — the frequency of the *Healthy* topic. To test whether informa-tion is present in the entire topic distribution, we use a supervised learning regime where we train a support vector machine with a ranking objective [9] to predict the infant's disease grade. In this supervised setting, we wanted to compare to features constructed by the fast Fourier transform, a commonly used feature processing scheme for time-series data. The FFT seems appropriate in this setting, since we observed earlier that words associated with the *Healthy* topic tend to exhibit higher frequencies. The frequencies of the resulting FFT coefficients span $1/v$ for $v \in \{1, \cdots, T\}$. Based on preliminary data analysis, we compute FFT features from a grid of periods between $4$ and $40$ minutes in increments of $4$ minutes. We use each of the FFT coefficients as a feature in the ranking SVM. To make for a fairer comparison, we estimate the parameters of TSTM without any supervision, using 6 topics; we use the proportions for each of the 6 topics as features of the time series in the ranking SVM.[3]. We report results averaged over 20 random folds with 50–50 train/test splits. The SVM tradeoff parameter $C$ was set using cross-validation. $\text{SVM}_{\text{TSTM}}$ and $\text{SVM}_{\text{FFT}}$ achieve mean performance (relative to maximum rank for each split) of $72.74\%$ and $63.5\%$ respectively (with a random order having a mean score of $50.0\%$). This suggests that the inferred topic distributions provide a useful feature representation scheme that can be employed within supervised objectives as an alternative or in addition to the FFT.

**Comparison to BP-AR-HMM:** As discussed in section 2, BP-AR-HMMs is the generative model for time series data that is most closely related to our work. For a visual comparison of the difference between TSTM and BP-AR-HMM, we used the BP-AR-HMM to infer word distributions over the same 30 infants shown in figure 3a. Since BP-AR-HMM is sampling in the space of binary matrices, it takes much longer to mix. Thus, due to computational resource constraints, we were only able to run BP-AR-HMMs on the first four days of data.

We experimented with different values of the parameters, but found that the model generally tends to encourage the generation of multiple unique, series-specific features (see example figure in supplementary materials). This behavior is not surprising, since the notion of individual series variation in BP-AR-HMMs is quite different from that in TSTMs: TSTMs, like topic models, encourage sharing by having all series utilize the same set of topics, but to different extents; by contrast, BP-AR-HMMs have some features that are shared and others that are explicitly series-specific. Thus, the two models are likely to be useful for fairly different types of exploratory data analysis.

We also performed a quantitative comparison regarding the utility of these features for a supervised learning task. We trained an SVM classifier, using leave-one-out cross-validation, to distinguish *Healthy* vs *Not healthy*, using the labels shown at the bottom of figure 3a. We compared the features derived from BP-AR-HMM from those inferred in an unsupervised manner using TSTM for a $4$ topic model. The SVM using the $4$ features learned by TSTM has an accuracy of $80\%$ while the SVM using the $43$ features inferred by BP-AR-HMM has an accuracy of $70\%$. Again, this is not surprising, since the fragmentation of the data across multiple individual features is bound to hurt performance, especially for such a small data set.

## 6 Discussion and Future work

In summary, the primary contribution of this paper is a new class of models for time-series data that emphasizes the modeling of instance specific variability while discovering population level characteristics. We demonstrate its use in a novel and useful application of modeling heterogeneous patient populations over time. Towards this end, we extended the existing topic modeling framework to also simultaneously discover the finite set of features (words) expressed in the data. We believe that TSTM provides a significant departure from current practices and a flexible tool for exploratory time series data analysis in novel domains. Furthermore, learned topic or word distributions can serve as features within supervised tasks. We demonstrated the utility of TSTMs on medical time series, but the framework is broadly applicable to other time-series applications.

Clinically, our work independently validates existing knowledge about heart rate dynamics containing information for characterizing health of a fetus [22] and predicting infection and death near onset time in premature infants[6]; furthermore, it suggests that information contained in the dynamics might be informative for a broader set of complications than infection and death.

---

[3]Due to the size of our corpus, the topic descriptions were inferred from only $30\%$ of the data and fixed at the $2000th$ iteration. Thereafter, we infer series-specific topic proportions based on these fixed settings of the model parameters.

Several extensions of LDA that model additional structure in the data [19, 20, 12] can add to the expressiveness of TSTMs. In particular, modeling disease composition over time (analogous to [19]) and disease evolution within a single patient (analogous to [20]) should provide interesting insight. We leave these next steps for future work.

## References

[1] Y. Bar-Shalom and T. E. Fortmann. *Tracking and Data Association*. Academic Press Professional, Inc., 1987.

[2] D. Blei, A. Ng, and M. Jordan. Latent Dirichlet allocation. In *J. Mach. Learn. Res.* 2003.

[3] E. Fox, E. Sudderth, M. Jordan, and A. Willsky. The sticky HDP-HMM: Bayesian nonparametric hidden Markov models with persistent states. Technical Report P-2777, MIT LIDS, 2007.

[4] E. Fox, E. Sudderth, M. Jordan, and A. Willsky. Nonparametric Bayesian learning of switching linear dynamical systems. In *NIPS*. 2008.

[5] E. Fox, E. Sudderth, M. Jordan, and A. Willsky. Sharing features among dynamical systems with Beta Processes. In *NIPS*. 2009.

[6] M. Griffin, D. Lake, E. Bissonette, F. Harrell, T. O'Shea, and J. Moorman. Heart rate characteristics: novel physiomarkers to predict neonatal infection and death. In *Pediatrics*. 2005.

[7] T. Griffiths and M. Steyvers. Finding scientific topics. In *PNAS*. 2004.

[8] H. Ishwaran and M. Zarepour. Exact and approximate sum-representation for the Dirichlet process. In *Canadian Journal of Statistics*. 2002.

[9] T. Joachims. Training linear SVMs in linear time. In *KDD*. 2006.

[10] K. Kurihara, M. Welling, and Y. W. Teh. Collapsed variational Dirichlet process mixture models. In *IJCAI*, 2007.

[11] H. Lee, Y. Largman, P. Pham, and A. Ng. Unsupervised feature learning for audio classification using convolutional deep belief networks. In *NIPS*, 2009.

[12] W. Li and A. McCallum. Pachinko Allocation: DAG-structured mixture models of topic correlations. In *ICML*, 2006.

[13] A. Mueen, E. Keogh, Q. Zhu, S. Cash, and B. Westover. Exact discovery of time series motifs. In *SDM*, 2009.

[14] J. Quinn, C. Williams, and N. McIntosh. Factorial switching linear dynamical systems applied to physiological condition monitoring. In *IEEE Trans. Pattern Analysis Machine Intelligence*, 2009.

[15] D. Ramage, D. Hall, R. Nallapati, and C. Manning. Labeled LDA: A supervised topic model for credit attribution in multi-labeled corpora. In *EMNLP*, 2009.

[16] D. Richardson, J. Gray, M. McCormick, K. Workman, and D. Goldmann. Score for neonatal acute physiology: a physiologic severity index for neonatal intensive care. In *Pediatrics*. 1993.

[17] J. Sethuraman. A constructive definition of Dirichlet priors. In *Statistics Sinica*. 1994.

[18] Y. W. Teh, M. Jordan, M. Beal, and D. Blei. Hierarchical Dirichlet processes. In *JASA*. 2006.

[19] C. Wang, D. Blei, and D. Heckerman. Continuous time dynamic topic models. In *UAI*. 2008.

[20] X. Wang and A. McCallum. Topics over Time: A non-Markov continuous time model of topical trends. In *KDD*, 2006.

[21] C. Williams, J. Quinn, and N. McIntosh. Factorial switching Kalman filters for condition monitoring in neonatal intensive care. In *NIPS*, 2005.

[22] K. Williams and F. Galerneau. Intrapartum fetal heart rate patterns in the prediction of neonatal acidemia. In *Am J Obstet Gynecol*. 2003.